\documentclass{article}

\usepackage{amsmath} 

\PassOptionsToPackage{
round,
semicolon,
}{natbib}

\usepackage[preprint]{neurips_2024}

\usepackage[utf8]{inputenc} 
\usepackage[T1]{fontenc}    
\usepackage{xcolor}         

\definecolor{darkblue}{HTML}{000080}
\usepackage[colorlinks, citecolor=darkblue, linkcolor=darkblue, urlcolor=darkblue]{hyperref}

\usepackage{url}            
\usepackage{booktabs}       
\usepackage{amsfonts}       
\usepackage{nicefrac}       
\usepackage{microtype}      
\usepackage{amsmath}
\usepackage{amssymb}
\usepackage{subfigure}
\usepackage{wrapfig}
\usepackage[all]{nowidow}
\usepackage{fp}
\usepackage{array}
\usepackage{geometry}
\usepackage{tabularx}
\usepackage{upgreek}
\usepackage{tcolorbox}
\usepackage{listings}
\usepackage[symbol]{footmisc}
\usepackage{placeins}

\definecolor{darkred}{HTML}{C23B22}
\definecolor{green}{HTML}{1cc650}
\definecolor{darkergreen}{HTML}{006400}

\definecolor{no_system}{rgb}{0.36, 0.54, 0.66}
\colorlet{prompt_specific}{darkgray!85}
\colorlet{baseline_color}{darkgray!80}

\colorlet{much_better}{green}
\colorlet{better}{green!65}
\colorlet{slightly_better}{green!35}
\colorlet{slightly_worse}{red!25}
\colorlet{worse}{red!50}
\colorlet{much_worse}{red}

\newcommand{\colornumber}[1]{%
  \FPeval{\result}{clip(#1)}%
  \ifdim \result pt < -1.99pt
    \textcolor{much_worse}{#1}%
  \else\ifdim \result pt < -0.99pt
    \textcolor{worse}{#1}%
  \else\ifdim \result pt < 0pt
    \textcolor{slightly_worse}{#1}%
  \else\ifdim \result pt < 0.99pt
    \textcolor{slightly_better}{#1}%
  \else\ifdim \result pt < 1.99pt
    \textcolor{better}{#1}%
  \else
    \textcolor{much_better}{#1}%
  \fi\fi\fi\fi\fi
}

{
  \begin{tcolorbox}[left=1.5mm, right=1.5mm, top=1.5mm, bottom=1.5mm]
    \raggedright
    \small
    \ifx\relax#1\relax\else
      \begin{center}
        {\normalsize \textbf{\color{black} #1}}
      \end{center}
    \fi
    \textcolor{black}{\textbf{Prompt:} {\texttt{#2}}} \\[2pt]
    \textcolor{darkergreen}{\textbf{Generation (no intervention):} {\texttt{#3}}} \\[2pt]
    \textcolor{darkred}{\textbf{Generation (intervention):} {\texttt{#4}}}
  \end{tcolorbox}
}{}

{
  \noindent \textcolor{black}{\textbf{Prompt:}} \textcolor{black}{\texttt{}} 
  \begingroup \color{black} \ttfamily
}{\endgroup \\[2pt]}

{
  \noindent \textcolor{darkergreen}{\textbf{Generation (no intervention):}} 
  \begingroup \color{darkergreen} \ttfamily
}{\endgroup \\[2pt]}

{
  \noindent \textcolor{darkred}{\textbf{Generation (intervention):}} 
  \begingroup \color{darkred} \ttfamily
}{\endgroup \\[2pt]}

{
  \begin{tcolorbox}[left=1.5mm, right=1.5mm, top=1.5mm, bottom=1.5mm]
    \raggedright
    \small
    \ifx\relax#1\relax\else
      \begin{center}
        {\normalsize \textbf{\color{black} #1}}
      \end{center}
    \fi
}{
  \end{tcolorbox}
}

\usepackage{titlesec}
\titlespacing*{\section}{0pt}{3.5ex plus 1ex minus .2ex}{2.3ex plus .2ex}

\usepackage{natbib}
\title{On Implications of Scaling Laws on Feature Superposition}

\author{
  Pavan Katta$^{\ast}$\\
}

\begin{document}

\maketitle

\renewcommand\footnoterule{\hrule width 2in height 0.4pt \vspace{6pt}}
\footnotetext[1]{Correspondence to \texttt{\href{mailto:pavanyellow@gmail.com}{pavanyellow@gmail.com}}}

\begin{abstract}
Using results from scaling laws, this theoretical note argues that the following two statements cannot be simultaneously true:
\begin{enumerate}
\item Superposition hypothesis where sparse features are linearly represented across a layer is a complete theory of feature representation.
\item Features are universal, meaning two models trained on the same data and achieving equal performance will learn identical features.
\end{enumerate}
\end{abstract}

\section{Introduction}

Scaling laws for language models give us a relation for a model's macroscopic properties such as cross entropy loss $L$, Amount of Data $D$ used and Number of non-embedding parameters $N$ in the model \citep{kaplan2020scaling}.
\begin{equation}
L(N,D) = \left[ \left( \frac{N_c}{N} \right)^{\frac{\alpha_N}{\alpha_D}} + \left( \frac{D_c}{D} \right)^{\alpha_D} \right]^{\alpha_D}
\end{equation}
where $N_c$, $D_c$, $\alpha_N$, and $\alpha_D$ are constants for a given task such as language modeling.

The scaling laws are not mere empirical observations and can be seen as predictive laws on limits of language model performance. During training of GPT-4, OpenAI was able to predict the final loss of GPT-4 early in the training process using scaling laws with high accuracy \citep{openai2024gpt4}.

An important detail is that the relation is expressed in terms of the number of parameters. It's natural to think of a model's computational capacity in terms of parameters, as they are the fundamental independent variables that the model can tune during learning. The amount of computation that a model performs in FLOPs for each input is also estimated to be $2N$ \citep{kaplan2020scaling}.

Let's compare this with Interpretability, where the representation of a feature is defined in terms of neurons or groups of neurons. At first glance, it might seem unnecessary to distinguish between computational capacity and feature representational capacity, as parameters are connections between neurons after all. However, we can change the number of neurons in a model while keeping the number of parameters constant. Kaplan et al. found that Transformer performance depends very weakly on the shape parameters $n_{layer}$ (number of layers), $n_{heads}$ (number of attention heads), and $d_{ff}$ (feedforward layer dimension) when we hold the total non-embedding parameter count $N$ fixed \citep{kaplan2020scaling}. The paper reports that the aspect ratio (the ratio of number of neurons per layer to the number of layers) can vary by more than an order of magnitude, with performance changing by less than 1

In this paper, we assume the above to be true and consider the number of parameters to be the true limiting factor, and we can achieve similar model performance for a range of aspect ratios. We then apply this as a postulate to the superposition hypothesis, our current best and successful theory of feature representation, and explore the implications.

The superposition hypothesis states that models can pack more features than the number of neurons they have \citep{elhage2022superposition}. There will be interference between the features as they can't be represented orthogonally, but when the features are sparse enough, the benefit of representing a feature outweighs the cost of interference. Concretely, given a layer of activations of $m$ neurons, we can decompose it linearly into activations of $n$ features, where $n > m$, as:
\begin{equation}
activation_{layer} = x_{f_1} W_{f_1} + x_{f_2} W_{f_2} + \cdots + x_{f_n} W_{f_n}
\end{equation}
where $activation_{layer}$ and $W_{f_i}$ are vectors of size $m$, and $x_{f_i}$ represents the magnitude of activation of the $i$-th feature. Sparsity means that for a given input, only a small fraction of features are active, which means $x_{f_i}$ is non-zero for only a few values of $i$.

\section{Case study on changing Aspect Ratio}

\begin{table}[t]
\caption{Macroscopic Properties of Transformer Models with Different Aspect Ratios but Equal Parameters}
\label{comparison-table}
\centering
\begin{tabular}{lcc}
\toprule
 & Model A & Model B \\
\midrule
Total Parameters & $d_{model}^2n_{layer}$ & $d_{model}^2n_{layer}$ \\[2pt]
Neurons per Layer & $d_{model}$ & $2d_{model}$ \\[2pt]
Number of Layers & $n_{layer}$ & $\frac{n_{layer}}{4}$ \\[2pt]
Total Number of Neurons & $d_{model}n_{layer}$ & $\frac{d_{model}n_{layer}}{2}$ \\[2pt]
Total Number of Features Learned & $F$ & $F$ \\[2pt]
Number of Features per Layer & $\frac{F}{n_{layer}}$ & $\frac{4F}{n_{layer}}$ \\[2pt]
Features per Neuron & $\frac{F}{d_{model}n_{layer}}$ & $\frac{2F}{d_{model}n_{layer}}$ \\
\bottomrule
\end{tabular}
\end{table}

Let's consider two Transformer models, Model A and Model B, having the same macroscopic properties. Both have an equal number of non-embedding parameters, are trained on the same dataset, and achieve similar loss according to scaling laws \citep{kaplan2020scaling}. However, their shape parameters differ. Using the same notation as Kaplan et al., let's denote the number of layers as $n_{layer}$, and number of neurons per layer as $d_{model}$. Model B has twice the number of neurons per layer compared to A. As the number of parameters is approximated by $d_{model}^2n_{layer}$, Model B must have $\dfrac{1}{4}$ the number of layers to maintain the same number of parameters as Model A. This means Model B has 8 times the aspect ratio ($\dfrac{d_{model}}{n_{layer}}$) of A which falls under the reported range in Kaplan et al.

The total number of neurons in a model is calculated by multiplying the number of neurons per layer by the number of layers. As a result, Model B has half the total number of neurons compared to Model A.

Now, let's apply the superposition hypothesis, which states that features can be linearly represented in each layer. Since both models achieve equal loss on the same dataset, it's reasonable to assume that they have learned the same features \citep{olah2020zoom}. Let's denote the total number of features learned by both models as $F$.

The above three paragraphs are summarized in Table \ref{comparison-table}.

The average number of features per neuron is calculated by dividing the number of features per layer by the number of neurons per layer. In Model B, this value is twice as high as in Model A, which means that Model B is effectively compressing twice as many features per neuron, in other words, there's a higher degree of superposition. However, superposition comes with a cost of interference between features, and a higher degree of superposition requires more sparsity.

Elhage et al.\citep{elhage2022superposition} show that, using lower bounds of compressed sensing \citep{ba2010lower}, if we want to recover \( n \) features compressed in \( m \) neurons (where \( n > m \)), the bound is 
\begin{equation}
m = \Omega(-n(1-S)\log(1-S)),
\end{equation}
where \( 1-S \) is the sparsity of the features. For example, if a feature is non-zero only 1 in 100 times, then \( 1-S \) equals 0.01. We can define the degree of superposition as 
\begin{equation}
\frac{n}{m} = \frac{1}{(1-S)\log(1-S)}
\end{equation}
which is a function of sparsity, inline with our theoretical understanding.

So Model B, with higher degree of superposition, should have sparser features compared to Model A. But, sparsity of a feature is a property of the data itself, and the same feature can't be sparser in Model B if both models are trained on the same data. This might suggest that they are not the same features, which breaks our initial assumption of two models learning the same features. So either our starting assumption of feature representation through superposition or feature universality needs revision. In the next section, we discuss how we might modify our assumptions.

\section{Discussion}

To recap, we started with the postulate that model performance is invariant over a wide range of aspect ratios and arrived at the inconsistency between superposition and feature universality. Though we framed the argument through the lens of superposition, the core issue is that the model's computational capacity is a function of parameters whereas the model's representational capacity is a function of total neurons.

A useful, though non-rigorous analogy, is to visualize a solid cylinder of radius $d_{model}$ and height $n_{layer}$. The volume (parameters) of the cylinder can be thought of as computational capacity whereas features are represented on the surface (neurons). We can change the aspect ratio of the cylinder while keeping the volume constant by stretching or squashing it. This changes the surface area accordingly. Though this analogy doesn't include sparsity, it captures the essentials of the argument in a simple way.

\begin{figure}[ht]
\centering
\includegraphics[width=0.9\columnwidth]{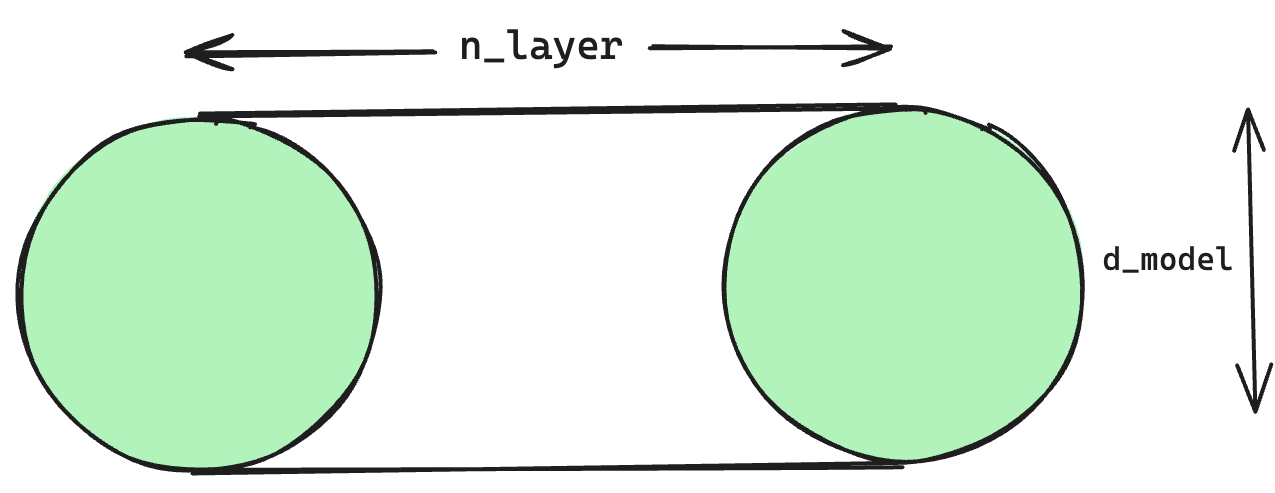}
\caption{A visual representation of the cylinder analogy, showing the relationship between the number of neurons per layer and the number of layers. The volume of the cylinder represents the total number of parameters, while the surface area corresponds to the total number of neurons available for feature representation.}
\label{cylinder-analogy}
\end{figure}

Coming to solutions, I do not have one that's consistent with scaling laws, superposition hypothesis, and feature universality, but will speculate on what a possible one might look like.

\subsection{Schemes of Compression Alternative to Superposition}

A crude and simple way to convert the total number of features into a function of parameters is to add a square term to compressed sensing bounds so it becomes $n = m^2.f(1-S)$. But this would require a completely new compression scheme compared to superposition. Methods such as Dictionary learning which disentangle features assuming superposition hypothesis have been successful for extracting interpretable features \citep{bricken2023monosemanticity}. So it's not ideal to ignore it, representation schemes whose first-order approximation looks like superposition might be more viable.

This isn't to say there's nothing we can improve on in the superposition hypothesis. Although dictionary learning features in Bricken et al. are much more mono-semantic than individual neurons, the lower activation levels in these features still look quite polysemantic.

\subsection{Cross Layer Superposition}

Previously, we used to look for features in a single neuron \citep{radford2017learning}, now we extended it to a group of neurons in a layer. A natural progression is to look for features localizing to neurons across multiple layers. But Model B from the above section, has half the number of neurons as A and the same inconsistencies would arise if features grow linearly on the number of neurons. Number of features represented across two or more layers by cross-layer superposition should grow superlinearly if Model B were to compensate for fewer neurons and still have the same representational capacity.

\section*{Acknowledgements}

I'm thankful to Jeffrey Wu and Tom McGrath for their helpful feedback on this topic. Thanks to Vinay Bantupalli for providing feedback on the draft. Earlier version of this work was supported by an Open Philanthropy research grant.

\bibliography{main.bib}
\bibliographystyle{plainnat}

\end{document}